%% file: iclr2027_conference.tex
\documentclass{article} 
\usepackage{iclr2027_conference,times}

\usepackage{hyperref}
\usepackage{url}
\usepackage{booktabs}
\usepackage{multirow}
\usepackage{makecell}
\usepackage{pifont}
\usepackage{graphicx}
\usepackage{array}
\usepackage{tabularx}
\usepackage{colortbl}

\usepackage[most]{tcolorbox}
\usepackage{xcolor}
\usepackage{enumitem}
\usepackage{float}

\newtcolorbox{promptbox}[1]{
  enhanced,
  breakable,
  colback=gray!3,
  colframe=black!80,
  boxrule=0.8pt,
  arc=2mm,
  left=4mm,
  right=4mm,
  top=5mm,
  bottom=4mm,
  before skip=6pt,
  after skip=10pt,
  fontupper=\small,
  title={#1},
  fonttitle=\bfseries,
  coltitle=white,
  attach boxed title to top left={
    xshift=4mm,
    yshift=-2mm
  },
  boxed title style={
    colback=black,
    colframe=black,
    arc=1mm,
    boxrule=0pt,
    left=2mm,
    right=2mm,
    top=1mm,
    bottom=1mm
  }
}

\title{DUDA-Bench: Benchmarking LLM Agents on Multimodal Data-Driven Urban Diagnosis}

\author{
Yizhi Song \quad
Hang Ni \quad
Weijia Zhang \quad
Hao Liu\thanks{Corresponding author.} \\
The Hong Kong University of Science and Technology (Guangzhou) \\
Guangzhou, China \\
\texttt{\{ysong531,hni017,wzhang411\}@connect.hkust-gz.edu.cn} \\
\texttt{liuh@hkust-gz.edu.cn}
}

\iclrfinalcopy

\usepackage{fancyhdr}
\renewcommand{\headrulewidth}{0pt}

\begin{document}

\maketitle
\pagestyle{plain}

\begin{abstract} Urban diagnosis integrates heterogeneous observations to identify urban problems, localize affected areas, and investigate contributing factors, informing evidence-based urban planning and management. However, its reliance on labor-intensive, case-specific expert workflows limits scalability and reuse, motivating the exploration of agent-based execution. To evaluate this capability, we introduce \textbf{DUDA-Bench}, a hierarchical and interactive benchmark that formalizes data-driven urban diagnosis as a multi-stage agent workflow. It comprises 86 atomic and 22 workflow tasks spanning four analytical stages, grounded in multimodal data from 12 cities covering five urban problem types. Evaluations of seven backbone models and five agent systems reveal a substantial gap between isolated analytical competence and end-to-end diagnosis, with system benefits varying across backbones. Trajectory analysis shows that unresolved evidence gaps propagate across stages, while successful recovery involves revising assumptions and actions using feedback. These findings highlight limitations in coordinating analytical capabilities across stages, particularly adaptive planning, evidence integration, and verification. More broadly, DUDA-Bench provides a framework for translating expert analytical workflows into hierarchical agent tasks and process-aware evaluation, supporting systematic assessment of end-to-end analytical capabilities. 
\end{abstract}

\section{Introduction}
\input{sections/1-introduction}

\section{Related Work}
\label{sec:related_work}
\input{sections/2-related-work}

\section{DUDA-Bench}
\label{sec:benchmark}
\input{sections/3-benchmark-design}

\section{Experiments}
\label{sec:experiments}
\input{sections/4-experiments}

\section{Conclusion}
\input{sections/5-conclusion}

\subsection*{AI use statement}
In this work, we used generative AI tools to translate and refine author-prepared text, assist literature discovery, and discuss and refine research ideas and methodology grounded in our review of prior work. The authors determined the substantive content and organization of the manuscript and manually checked retrieved publications and references against their original sources. Codex and Claude assisted with code implementation and debugging. LLMs were also used as evaluated agents and for AI-assisted judging, as described in the evaluation protocol(Section~\ref{sec:evaluation}). We did not use generative AI to synthesize the underlying urban observations, which originate from real-world data sources documented in Appendix~\ref{app:dataset-statistics}; mathematical proof assistance is not applicable to this work. The authors reviewed all AI-assisted work and take responsibility for the final text, code, analyses, claims, and other artifacts.

\subsection*{Ethics statement}

DUDA-Bench uses real-world urban data from publicly accessible sources documented in Appendix~\ref{app:dataset-statistics}. Our dataset release will comply with the applicable source licenses and terms of use, preserve required attribution, and redistribute only materials for which redistribution is permitted; otherwise, we will provide source references and instructions for authorized access. Released materials will be reviewed for privacy risks, with potentially identifying content excluded or appropriately protected. Geographic coverage and source-data biases may affect benchmark results and limit their generalization. DUDA-Bench is intended for research evaluation, and agent-generated diagnoses should not be used for consequential planning decisions without expert review and local validation.

\subsection*{Reproducibility statement}

The benchmark construction section describes the task hierarchy, multimodal data organization, interactive environment, and evaluation protocol, while the experimental setup specifies the evaluated models, agent systems, and execution settings. Appendix~\ref{app:dataset-statistics} documents the data sources and dataset composition. Upon acceptance, we will publicly release the benchmark code and dataset, subject to the source-specific redistribution conditions described in the Ethics statement. The release will include data collection and preprocessing scripts, task definitions, evaluation code, experiment configurations, and a README with setup and execution instructions, supporting reproduction of the experiments and extension of the data collection pipeline to additional cities.

\bibliography{iclr2027_conference}
\bibliographystyle{iclr2027_conference}

\clearpage
\appendix

\input{sections/appendix}

\end{document}

%% file: sections/1-introduction.tex
\newcommand{\cmark}{\ding{51}}
\newcommand{\xmark}{\ding{55}}
\newcommand{\pmark}{$\triangle$}
\newcommand{\na}{--}

Urban diagnosis is the systematic assessment of urban conditions to identify problems and investigate their underlying causes, thereby informing urban planning decisions~\citep{ye2021community}. Data-driven studies have examined urban sprawl, heat islands, and accessibility using heterogeneous spatial, temporal, and socioeconomic evidence~\citep{SHRESTHA2012522,XIE2018350,WEI2024103875,ijgi14070247}. These analyses rely on experts to select evidence and iteratively refine interpretations, making them labor-intensive to reproduce across regions and problem types. LLM agents can reason over multiple steps and interact with external tools~\citep{yao2023react}, offering new opportunities for data-driven urban diagnosis. This motivates evaluating their ability to carry out complete diagnostic workflows and identifying the capabilities that limit success.

Assessing these capabilities requires evaluation settings that capture the analytical demands of urban diagnosis. Existing urban and geospatial benchmarks assess interactive decision making, multimodal understanding, and executable spatial analysis~\citep{10.1145/3711896.3737375,Feng_2025_ICCV,yu2026geoagentbenchdynamicexecutionbenchmark}, while general and scientific agent benchmarks evaluate multi-step tool use and trajectory-level execution~\citep{wang2026gta2benchmarkinggeneraltool,shen2026sciagentgym,liu2026towards}. These efforts provide relevant foundations, but leave open how effectively agents can sustain an evidence-driven diagnostic process across heterogeneous urban observations(Table~\ref{tab:benchmark_positioning}). Unlike a predefined operation such as indicator estimation, open-ended diagnosis requires agents to determine what evidence is needed, acquire and integrate observations, and refine conclusions through interaction, as illustrated by the urban-heat example in Fig.~\ref{fig:motivation} (Appendix~\ref{app:benchmark-construction}). Evaluating this process requires examining both diagnostic outcomes and intermediate analytical progress. The central challenge is to understand how individual analytical capabilities translate into coherent end-to-end diagnosis, and where their coordination breaks down.

To address this evaluation gap, we introduce \textbf{DUDA-Bench}, a benchmark for \textbf{data-driven urban diagnosis} that combines heterogeneous urban data, adaptive analytical workflows, and outcome- and process-level evaluation. It comprises two complementary task types: \textbf{workflow tasks} assess complete diagnostic processes spanning problem framing, urban profiling, diagnostic reasoning, and validation and synthesis, while \textbf{atomic tasks} assess individual analytical operations extracted from these workflows (Fig.~\ref{fig:DUDA_overview}). Workflow tasks extend predefined urban operations to diagnosing what the problem is, where it occurs, and what evidence supports its explanation through iterative analysis (Fig.~\ref{fig:motivation}). Together, the two task types distinguish competence on individual operations from the ability to coordinate them into a complete diagnosis.

DUDA-Bench integrates three components: a \textbf{hierarchical task taxonomy}~(\S\ref{sec:DUDA_task}) linking atomic operations to workflow tasks, a \textbf{multimodal urban case dataset}~(\S\ref{sec:DUDA_data}) providing standardized data across cities and problem types, and an \textbf{agentic environment}~(\S\ref{sec:DUDA_env}) supporting data inspection, domain tools, reusable skills, and code execution under multiple agent systems. Hierarchical outcome evaluation measures diagnostic success, while stage-aligned checkpoints assess intermediate analytical progress. Together, these components enable controlled comparisons and help locate where evidence acquisition, reasoning, and execution succeed or fail.

\input{tables_tex/tab1_benchmark_position}
We evaluate seven backbone models under a shared execution framework and compare five agent systems on two backbones, complemented by trajectory analysis and component ablations. The results reveal a substantial capability gap between isolated analytical operations and end-to-end diagnosis, with agent-system benefits varying across backbones. Tracing workflow execution shows how this gap develops: unresolved evidence gaps persist into downstream reasoning, whereas successful recovery involves revising assumptions and actions in response to feedback. Together, these findings highlight limitations in coordinating analytical capabilities across stages, particularly adaptive planning, evidence integration, and verification. Component ablations further show that these capabilities depend on access to computational and domain-specific support during execution.

Accordingly, our main contributions are summarized as follows:
\begin{itemize}
    \item We formulate \textbf{data-driven urban diagnosis} as an open-ended agentic task and establish a hierarchical task design linking individual analytical operations to complete diagnostic workflows. This formulation enables evaluation of the gap between component capabilities and end-to-end task completion.
    \item We build \textbf{DUDA-Bench} with standardized multi-modal urban cases, an interactive environment supporting domain tools, reusable skills, and code execution, and a protocol combining outcome assessment with stage-aligned checkpoints. These resources support evaluation across regions, problem types, and agent systems.
    \item We systematically evaluate backbone models and agent systems, revealing a substantial gap between atomic competence and workflow-level diagnosis. Trajectory analysis and component ablations highlight limitations in cross-stage capability coordination and the role of interactive execution in supporting diagnostic performance.
\end{itemize}

\input{figures_tex/fig_overview}

%% file: tables_tex/tab1_benchmark_position.tex
\begin{table*}[t]
    \centering
    \caption{
        Positioning comparison of DUDA-Bench with representative urban,
        geospatial, and Earth-observation benchmarks.
    }
    \label{tab:benchmark_positioning}
    \setlength{\tabcolsep}{2pt}
    \renewcommand{\arraystretch}{1.06}

    \resizebox{\textwidth}{!}{
    \begin{tabular}{@{}ll*{11}{c}@{}}
        \toprule

        \multirow{2}{*}{\textbf{Category}}
        & \multirow{2}{*}{\textbf{Benchmark}}
        & \multicolumn{6}{c}{\textbf{Data Coverage}}
        & \multicolumn{3}{c}{\textbf{Agentic Workflow Tasks}}
        & \multicolumn{2}{c}{\textbf{Evaluation}} \\

        \cmidrule(lr){3-8}
        \cmidrule(lr){9-11}
        \cmidrule(lr){12-13}

        &
        & \makecell{\textbf{Map}\\\textbf{Data}}
        & \makecell{\textbf{Urban}\\\textbf{Indicators}}
        & \makecell{\textbf{Remote}\\\textbf{Sensing}}
        & \makecell{\textbf{Street}\\\textbf{View}}
        & \makecell{\textbf{Temporal}\\\textbf{Data}}
        & \textbf{Cities}
        & \makecell{\textbf{Interactive}\\\textbf{Execution}}
        & \makecell{\textbf{Multi-stage}\\\textbf{Workflow}}
        & \textbf{Planning}
        & \textbf{Outcome}
        & \makecell{\textbf{Process-}\\\textbf{aware}} \\

        \midrule

        \multirow{3}{*}{\makecell[l]{\textbf{Urban-agent}\\\textbf{benchmarks}}}
        & \makecell[l]{CityBench\\{\scriptsize\cite{10.1145/3711896.3737375}}}
        & \cmark & \cmark & \cmark & \cmark & \cmark
        & 13
        & \cmark & \cmark & \xmark
        & \cmark & \xmark \\

        & \makecell[l]{CityEQA\\{\scriptsize\cite{zhao-etal-2025-cityeqa}}}
        & \xmark & \xmark & \xmark & \cmark & \xmark
        & --
        & \cmark & \cmark & \cmark
        & \cmark & \xmark \\

        & \makecell[l]{USTBench\\{\scriptsize\cite{lai2026ustbench}}}
        & \cmark & \cmark & \xmark & \xmark & \cmark
        & --
        & \cmark & \cmark & \xmark
        & \cmark & \cmark \\

        \midrule

        \multirow{4}{*}{\makecell[l]{\textbf{Urban-multimodal}\\\textbf{benchmarks}}}
        & \makecell[l]{UrBench\\{\scriptsize\cite{10.1609/aaai.v39i10.33163}}}
        & \xmark & \xmark & \cmark & \cmark & \xmark
        & 11
        & \xmark & \xmark & \xmark
        & \cmark & \xmark \\

        & \makecell[l]{DynamicVL\\{\scriptsize\cite{xuan2026dynamicvl}}}
        & \xmark & \cmark & \cmark & \xmark & \cmark
        & 42
        & \xmark & \xmark & \xmark
        & \cmark & \xmark \\

        & \makecell[l]{UBench\\{\scriptsize\cite{Feng_2025_ICCV}}}
        & \cmark & \xmark & \cmark & \cmark & \xmark
        & 3
        & \cmark & \cmark & \xmark
        & \cmark & \xmark \\

        & \makecell[l]{CityLens\\{\scriptsize\cite{liu2026citylens}}}
        & \xmark & \cmark & \cmark & \cmark & \xmark
        & 17
        & \xmark & \xmark & \xmark
        & \cmark & \xmark \\

        \midrule

        \multirow{3}{*}{\makecell[l]{\textbf{Geospatial}\\\textbf{benchmarks}}}
        & \makecell[l]{GEOBench-VLM\\{\scriptsize\cite{Danish_2025_ICCV}}}
        & \xmark & \xmark & \cmark & \xmark & \cmark
        & --
        & \xmark & \xmark & \xmark
        & \cmark & \xmark \\

        & \makecell[l]{Earth-Agent\\{\scriptsize\cite{feng2026earthagent}}}
        & \xmark & \cmark & \cmark & \xmark & \cmark
        & --
        & \cmark & \cmark & \cmark
        & \cmark & \cmark \\

        & \makecell[l]{GeoAgentBench\\{\scriptsize\cite{yu2026geoagentbenchdynamicexecutionbenchmark}}}
        & \cmark & \xmark & \xmark & \xmark & \xmark
        & --
        & \cmark & \cmark & \cmark
        & \cmark & \cmark \\

        \midrule

        \textbf{Ours}
        & \textbf{DUDA-Bench}
        & \cmark & \cmark & \cmark & \cmark & \cmark
        & \textbf{12}
        & \cmark & \cmark & \cmark
        & \cmark & \cmark \\

        \bottomrule
    \end{tabular}
    }

    \begin{minipage}{\textwidth}
        \footnotesize
        \textit{Note.}
        Multi-temporal Data denotes repeated real-world observations over time.
        Planning denotes adaptive task decomposition and execution-path revision;
        Process-aware denotes evaluation of intermediate analytical progress.
        \cmark: supported; \xmark: not a primary component; --: not directly comparable.
    \end{minipage}
\end{table*}

%% file: figures_tex/fig_overview.tex
\begin{figure*}[t]
    \centering
    \includegraphics[width=\textwidth]{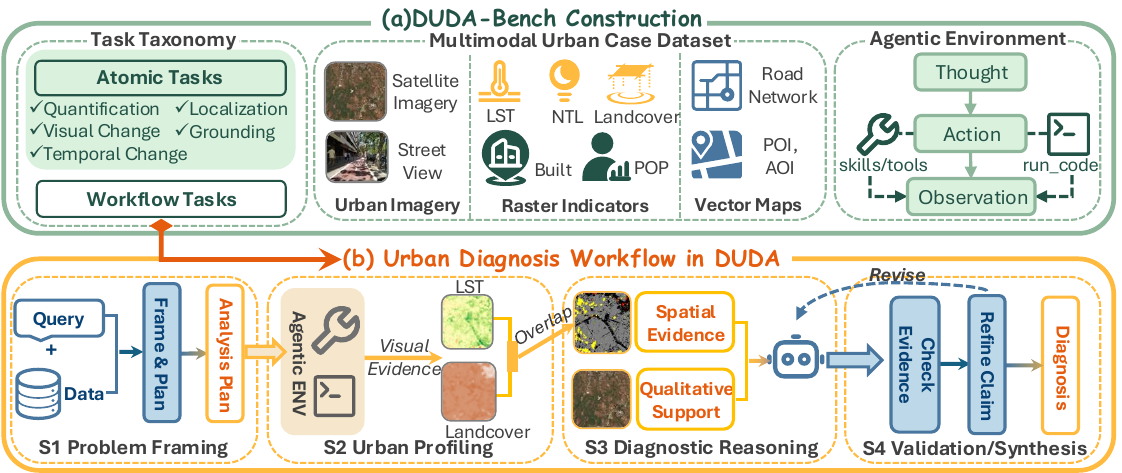}
    \caption{Overview of DUDA-Bench. The benchmark connects a hierarchical task taxonomy and multimodal urban case dataset with an interactive agent environment, and formulates urban diagnosis as a four-stage analytical workflow.}
    \label{fig:DUDA_overview}
\end{figure*}

%% file: sections/2-related-work.tex
\paragraph{Data-Driven Urban Diagnosis.} Urban problem diagnosis provides an analytical basis for understanding urban conditions and informing interventions~\citep{su10093306,BIBRI2021100047,FOSTER201692}. Existing data-driven studies typically address individual problems in specific regions, including urban sprawl~\citep{SHRESTHA2012522}, shrinkage~\citep{XIE2018350}, heat islands~\citep{WEI2024103875}, and accessibility~\citep{ijgi14070247}. These studies establish the conceptual and empirical foundations of urban diagnosis, but do not formulate it as a unified task across problem types, regions, and heterogeneous data sources. Building on this foundation, DUDA-Bench formalizes data-driven urban diagnosis as a common, evaluable agent task.

\paragraph{Urban and Agent Benchmarks.} Existing urban benchmarks evaluate complementary capabilities, including interactive urban agents~\citep{10.1145/3711896.3737375,zhao-etal-2025-cityeqa,lai2026ustbench}, multimodal urban understanding~\citep{10.1609/aaai.v39i10.33163,Feng_2025_ICCV,xuan2026dynamicvl,liu2026citylens}, and executable geospatial or Earth-observation analysis~\citep{Danish_2025_ICCV,feng2026earthagent,yu2026geoagentbenchdynamicexecutionbenchmark}. In parallel, general and scientific agent benchmarks increasingly evaluate multi-step execution and trajectories rather than only final answers~\citep{yao2025taubench,wang2026gta2benchmarkinggeneraltool,he2026trajectbencha,shen2026sciagentgym,liu2026towards,varambally2026zephyrus}. DUDA-Bench builds on these complementary directions by firstly formulating and benchmarking \textbf{data-driven urban diagnosis} as a unified agentic task, requiring agents to integrate heterogeneous urban evidence, execute multi-stage analyses interactively, and be evaluated at both outcome and process levels.

%% file: sections/3-benchmark-design.tex
\textbf{DUDA-Bench} is a hierarchical and interactive benchmark for data-driven urban diagnosis, built around three components: a task taxonomy, a multimodal urban dataset, and an agentic environment. Construction proceeds through four stages. \textbf{Literature-Grounded Task Construction} translates expert diagnostic workflows into workflow tasks and constituent atomic operations. \textbf{Multimodal Data Collection and Organization} grounds these tasks in observations from 12 cities covering five urban problem types over a ten-year period. \textbf{Interactive Environment Construction} makes these data accessible through tools, skills, and code execution, enabling agents to carry out the tasks through traceable multi-turn interaction. Finally, \textbf{Hierarchical Evaluation} combines task-outcome assessment with stage-aligned checkpoints to evaluate both individual analytical capabilities and the execution of complete diagnostic workflows.

\subsection{Task Formulation}
\label{sec:task_formulation}
In DUDA-Bench, we formulate data-driven urban diagnosis as a multi-step, tool-assisted task $\mathcal{T}=(q,\mathcal{D},\mathcal{E})$, where $q$ specifies the diagnostic objective and spatiotemporal scope, $\mathcal{D}$ denotes the available multimodal urban data, and $\mathcal{E}$ is an executable environment providing data access and analytical tools. Given $\mathcal{T}$, an agent iteratively gathers and analyzes evidence, with intermediate results informing subsequent actions and revisions. The goal is to produce an evidence-grounded diagnosis that identifies the urban problem, localizes affected areas, and explains contributing factors.

\subsection{Task Taxonomy}
\label{sec:DUDA_task}
\textbf{Task construction.} We construct tasks in two steps. First, two urban-domain experts, each with over five years of research experience, manually select and review urban diagnosis studies~\citep{su10093306,BIBRI2021100047,FOSTER201692,SHRESTHA2012522,XIE2018350,WEI2024103875,ijgi14070247}. These studies cover five typical scenarios grouped into three categories: urban morphology (urban sprawl and shrinkage), eco-climatic resilience (urban heat islands and green-space fragmentation), and service accessibility (accessibility inequality). The experts translate the studies' diagnostic objectives into workflow tasks and extract their reported conclusions as reference ground truth. Second, we extract individual analytical operations from these workflows and formulate atomic tasks according to five capability types: indicator quantification, temporal change analysis, spatial problem localization, visual change interpretation, and evidence-to-metric grounding. Atomic reference outputs are computed using the benchmark dataset and manually checked. Tasks at both levels are reviewed for data availability and evaluation feasibility.

\textbf{Atomic-to-Workflow Task Taxonomy.} The resulting taxonomy comprises \textbf{22 workflow tasks} and \textbf{86 atomic tasks}, with their composition and distribution summarized in Table~\ref{tab:task_statistics} in Appendix~\ref{app:task}. Workflow tasks follow four analytical stages (Fig.~\ref{fig:DUDA_overview}): \textbf{(S1) Problem Framing} defines the target issue and spatiotemporal scope; \textbf{(S2) Urban Profiling} gathers multimodal observations and computes indicators; \textbf{(S3) Diagnostic Reasoning} integrates evidence to identify problems and interpret possible mechanisms; and \textbf{(S4) Validation and Synthesis} checks evidence, revises claims, and synthesizes the diagnosis. Across these scenarios, workflow tasks are grouped into three complexity levels according to the number of urban issues within the same geographic unit: single-issue, dual-issue, and multi-issue tasks involve one, two, and three or more issues, respectively. Atomic tasks assess individual operations within this structure, while workflow tasks require agents to select and coordinate these capabilities through open-ended, multi-turn analysis. This paired design supports assessment of both individual analytical competence and end-to-end diagnostic performance. Further construction details are provided in Appendix~\ref{app:task}.

\subsection{Multimodal Urban Case Dataset}
\label{sec:DUDA_data}

\textbf{Dataset scope.} We construct a multimodal urban dataset to support cross-region and cross-problem evaluation of data-driven urban diagnosis. Rather than combining existing benchmark datasets, we collect observations directly from original data providers and standardize them under consistent spatial scopes, temporal organization, and annotation criteria. The dataset covers two spatial levels: city-level observations capture broad urban change, while study-region observations focus on case-specific areas identified from the literature and support finer-grained localization. It contains data for \textbf{12 cities} and \textbf{22 study regions} over \textbf{2015--2024}, spanning North America, Europe, and Asia. Totally, our dataset contains 7,071 data assets and 1,507,693 data records. Dataset statistics and composition are summarized in Table~\ref{tab:dataset_statistics} in Appendix~\ref{app:dataset-statistics}.

\textbf{Data collection and standardization.} The city pool and data requirements are grounded in empirical studies of the five target urban problem types~\citep{SHRESTHA2012522,XIE2018350,WEI2024103875,ijgi14070247}. We assemble complementary urban data from eight sources: OSM, WorldPop, GHSL, VIIRS, Landsat 8/9, ESA WorldCover, Sentinel-2 L2A, and Mapillary. Each source is clipped or queried for the predefined city or study region and organized with harmonized spatial and temporal metadata. Table~\ref{tab:dataset_statistics} summarizes the dataset's coverage, composition, and scale, while Table~\ref{tab:data_source_details} details source contents and access channels. Both of them appear in Appendix~\ref{app:dataset-statistics}.

\subsection{Agentic Environment}
\label{sec:DUDA_env}
\textbf{Interactive execution.} DUDA provides a unified interactive environment for urban-diagnosis agents implemented with different systems, including Direct prompting, Plan-and-Execute~\citep{wang2023planandsolvepromptingimprovingzeroshot}, ReAct~\citep{yao2023react}, MAF-Magentic~\citep{fourney2024magenticonegeneralistmultiagentsolving}, and LAMBDA~\citep{Sun_2025}. The environment supports multi-turn interaction through standardized interfaces to urban data, domain-specific tools, reusable skills, and a code sandbox. Agents can iteratively inspect data, compute indicators, validate intermediate results, and revise their analysis based on accumulated evidence. All systems share the same task inputs and tool interfaces, supporting controlled comparisons of backbone models and agent systems while allowing each system to follow its own execution strategy.

\textbf{Domain tools, skills, and data access.} The environment provides 7 domain skills and 8 tool families to support urban diagnosis. Skills offer reusable guidance on data selection, indicator definition, and spatiotemporal comparison, while tools enable data retrieval, computation, visualization, and code execution. Through these interfaces, agents access evidence as \textit{numerical values}, \textit{visual evidence}, and \textit{textual descriptions}. Visual evidence includes both source imagery and visualizations generated from raster data, allowing agents to combine quantitative analysis with visual interpretation. For analyses beyond the predefined tools, agents can use \texttt{run\_code} to process data and implement custom computations in the sandbox. The complete mapping between domain skills, tool families, and executable operations is provided in Appendix~\ref{app:domain_skills}.

\subsection{Evaluation Protocol}
\label{sec:evaluation}

\textbf{Outcome Evaluation.} We evaluate atomic operations and complete diagnostic workflows separately. Atomic outputs are checked by task-specific validators for numerical accuracy, spatial agreement, directional correctness, or structural validity. A task passes only if its output meets the specified criterion. Workflow evaluation requires more than a correct issue label: agents must also locate the problem and provide quantitative support. Following urban-analysis practices that characterize problems through their definition, spatial pattern, and diagnostic indicators~\citep{ZHANG2023104776,LIU2023116303,DEILAMI201830}, we assess three aspects: \textbf{Recognition} ($A$), identification of the target issue family; \textbf{Localization} ($B$), coverage of the reference locations; and \textbf{Indicator F1} ($C$), precision and recall of the reported diagnostic evidence.

For a single run on a workflow target, let $f$ be the target issue family and $\widehat{\mathcal F}$ the predicted families. Localization and evidence are evaluated using only predictions associated with $f$. We compute:
\begin{align}
A &= \mathbf{1}\{f \in \widehat{\mathcal F}\}, \\
B &= A \cdot \operatorname{Coverage}(\widehat{\mathcal L},\mathcal L),\\
C &= A \cdot \frac{2|\widehat{\mathcal E}\cap\mathcal E|}{|\widehat{\mathcal E}|+|\mathcal E|}, \\
S &= A \cdot \frac{B+C}{2}.
\end{align}
Here, $\mathcal L$ and $\widehat{\mathcal L}$ denote the reference and predicted locations associated with the target issue. $\operatorname{Coverage}\in[0,1]$ measures the proportion of reference locations covered by the prediction, using spatial overlap or matched areas of interest. Where an area constraint applies, predictions covering more than 25\% of the city's bounding box receive zero localization credit, preventing agents from gaining credit through overly broad spatial predictions. $\mathcal E$ and $\widehat{\mathcal E}$ denote the required and reported diagnostic evidence sets. Incorrect issue recognition yields zero localization, indicator, and overall scores.

We first average scores across repeated runs of each task or target, then macro-average across atomic tasks or workflow targets:
\begin{equation}
\mathrm{Atomic\ PR}=\frac{1}{N_a}\sum_{i=1}^{N_a}\bar p_i,
\qquad
\mathrm{Workflow\ Score}=\frac{1}{N_w}\sum_{i=1}^{N_w}\bar S_i,
\end{equation}
where PR denotes pass rate, $\bar p_i$ is the mean binary pass outcome for atomic task $i$, $\bar S_i$ is the mean continuous score for workflow target $i$, and $N_a$ and $N_w$ are their respective counts.

\textbf{Checkpoint-Driven Trajectory Evaluation.} Outcome metrics indicate whether an agent completes the diagnosis but not where the execution diverges. Inspired by GTA-2~\citep{wang2026gta2benchmarkinggeneraltool}, we therefore decompose each workflow into 14 verifiable checkpoints aligned with the four analytical stages. Each checkpoint specifies an analytical sub-goal rather than a prescribed tool sequence, and GLM-5.3~\citep{glm5team2026glm5vibecodingagentic} judges the observable trajectory as \textit{pass}, \textit{partial}, or \textit{fail}. Representative checkpoint results are reported in Fig.~\ref{fig:exp}(B4). We further validate the judge on 56 randomly sampled ReAct trajectories (716 checkpoint verdicts): two independent annotators achieve 93.6\% agreement, while GLM-5.3 reaches 90.4\% accuracy against adjudicated human labels. Additional verification details are provided in Appendix~\ref{app:human-verification}.

%% file: sections/4-experiments.tex
We conduct experiments to address three research questions(RQs) that progressively examine overall performance, execution dynamics, and capability bottlenecks in multimodal urban diagnosis. \textbf{RQ1:} How do current LLM/MLLM agents perform on multimodal data-driven urban diagnosis, and how does performance vary across backbone models and agent execution strategies? \textbf{RQ2:} How do failures and recoveries unfold over multi-stage diagnostic workflows, and what trajectory-level behaviors distinguish successful from unsuccessful executions? \textbf{RQ3:} Which analytical capabilities are most critical to successful end-to-end urban diagnosis, and how do weaknesses in these capabilities manifest in workflow-level performance? Our experiments follow this progression, beginning with overall performance and controlled backbone and agent-system comparisons, then examining variation across benchmark factors, and finally tracing workflow executions to identify failure patterns and the capabilities underlying them.

\subsection{Experimental Setup}

\paragraph{Models and Agent Systems.} We organize our experiments into three complementary settings to disentangle the effects of backbone capability, agent-system design, and individual execution components. First, for the \textit{backbone comparison}, we evaluate seven representative frontier and open-weight models: GPT-5.6-Terra~\citep{openai2026gpt56}, GPT-5.6-Sol~\citep{openai2026gpt56}, Gemini-3.7-Flash~\citep{google2026gemini37flash}, Claude-Sonnet-5~\citep{anthropic2026sonnet5}, Kimi-K3~\citep{kimiteam2026kimik3openfrontier}, Qwen3.8-Max and Qwen3.8-27B~\citep{qwen3.8}. To isolate backbone capability from agent scaffolding, all models are evaluated under the same ReAct-style execution framework~\citep{yao2023react}, which supports multi-turn reasoning and interaction with a code sandbox, domain-specific tools, and reusable skills. Each backbone is evaluated on the full benchmark comprising 22 workflow tasks and 86 atomic tasks. Second, for the \textit{agent-system comparison}, we fix the backbone to Qwen3.8-Max and Qwen3.8-27B and compare five execution paradigms: Direct prompting, Plan-and-Execute, ReAct, MAF-Magentic based on the Magentic multi-agent orchestration paradigm, and LAMBDA, a specialized multi-agent system for data analysis. This comparison examines how different levels of planning, iterative environment interaction, multi-agent coordination, and domain-specific agent design affect urban diagnostic performance. Finally, we conduct a \textit{component ablation} using Qwen3.8-Max with ReAct as the base configuration, comparing the full system against three variants that remove code execution (\textit{w/o code}), skill routing (\textit{w/o skill-routing}), or domain tools (\textit{w/o tools}), respectively, to quantify the contribution of each execution component.

\textbf{Metrics.}
We report \textbf{Atomic PR} for atomic-task completion and four workflow-level outcome metrics: \textbf{Recognition}, \textbf{Localization}, \textbf{Indicator F1}, and the aggregated \textbf{Workflow Score}, as defined in Section~\ref{sec:evaluation}. To characterize execution efficiency, we additionally report the average number of agent turns (\textbf{Avg. Turns}) and total token usage in thousands (\textbf{Tokens (K)}) per workflow episode. Higher values are better for all outcome metrics, whereas lower Avg. Turns and Tokens indicate greater efficiency.

\input{tables_tex/tab_main_results}
\input{tables_tex/tab_agent_comparison}
\input{figures_tex/fig_exp}

\subsection{Overall Performance and Agent System Effects (RQ1)}
\label{sec:overall_results}

\textbf{End-to-end diagnosis remains substantially harder than isolated analytical tasks.} Table~\ref{tab:main_backbone_results} and Fig.~\ref{fig:exp}(B2) show that Qwen3.8-Max achieves the highest Workflow Score (0.277), while the remaining models range from 0.071 to 0.182. By contrast, Atomic PR varies within a narrower range of 0.419--0.558. Notably, Qwen3.8-Max, GPT-5.6-Sol, and Qwen3.8-27B all obtain an Atomic PR of 0.500 but diverge markedly at the workflow level. This suggests that competence on individual operations is insufficient for completing a multi-stage diagnostic process. Table~\ref{tab:agent_system_results} further shows that this gap depends on the execution system: Qwen3.8-Max performs best with ReAct, whereas Qwen3.8-27B performs best with LAMBDA.

\textbf{Tool-use strategies expose different evidence bottlenecks.} Execution traces reveal three broad styles: code-dominant behavior for Gemini-3.7-Flash and Qwen3.8-27B, high-level tool reliance for Claude-Sonnet-5 and GPT-5.6-Sol, and a more mixed strategy for Qwen3.8-Max. Code enables finer-grained spatial comparisons and specialized indicators, but introduces more execution failures, while high-level tools are more stable but tend to remain at aggregate statistics. Visual tools are rarely used across models, consistent with the weak visual-change capability in Fig.~\ref{fig:exp}(B6). Importantly, code frequency alone does not predict success: Gemini-3.7-Flash and Qwen3.8-27B are both code-heavy, yet their Workflow Scores differ substantially. What matters is whether tool use produces evidence that directly supports the diagnostic objective.

\textbf{System effects are backbone dependent, and higher cost does not imply better performance.} As shown in Table~\ref{tab:agent_system_results}, the relative advantage of Qwen3.8-Max over Qwen3.8-27B changes across systems and even reverses under LAMBDA, indicating a clear backbone--system interaction. Efficiency exhibits the same non-monotonic pattern. Table~\ref{tab:main_backbone_results} reports 9.9--20.3 turns and 177.9K--322.3K tokens per workflow, while Fig.~\ref{fig:exp}(B3) shows that Qwen3.8-Max achieves the strongest cost-normalized performance. Qwen3.8-27B consumes the largest interaction budget without attaining the highest Workflow Score. Effective diagnosis therefore depends more on how computation is allocated to evidence acquisition and analysis than on interaction budget alone.

\subsection{Failure Propagation and Recovery Patterns (RQ2)}
\label{sec:trajectory_analysis}

\textbf{Early evidence gaps tend to persist into downstream diagnosis.} Figure~\ref{fig:exp}(B4) shows that checkpoint failures are not confined to the final reasoning stage: agents often formulate plausible analysis plans but fail to convert required indicators into usable observations. This gap then propagates from profiling to diagnosis, producing trajectories in which a missing spatial or quantitative analysis is later reported as insufficient evidence rather than actively resolved. For example, GPT-5.6-Sol can identify the relevant modalities while still omitting discriminative indicators, whereas Qwen3.8-Max more consistently carries intermediate evidence into mechanism-level reasoning. The central difficulty is therefore not merely planning what to inspect, but maintaining alignment between the target problem, the evidence collected, and the downstream diagnostic claim.

\textbf{Successful execution does not guarantee valid evidence.} Fig.~\ref{fig:exp}(B5) attributes checkpoint failures to execution, tool selection, evidence use, planning, and validation. Runtime errors make failures explicit and can guide recovery, but analytically invalid outputs may go unnoticed. For example, Qwen3.8-27B sometimes uses inappropriate land-cover categories or produces grid statistics dominated by missing values despite successful execution. Without checks on category meaning, coordinate systems, and data validity, these outputs can mislead subsequent analysis. Repeated calls can still be productive when failures reveal new information about the data. Verification must therefore assess whether outputs support the analysis, beyond whether tools execute successfully.

\textbf{Successful recovery involves revising the analysis in response to errors.} Two trajectories illustrate this pattern. In a successful Qwen3.8-27B heat-analysis run, errors prompt the agent to check how raster coordinates map to geographic locations, which pixels contain valid data, and which years are available. These checks guide subsequent analysis and allow the agent to complete a localized comparison. In an unsuccessful sprawl run with a similar number of turns, the agent repeatedly encounters alignment and coding errors but continues without revising its assumptions, ultimately failing to produce reliable spatial evidence. Table~\ref{tab:agent_system_results} likewise shows that additional interaction opportunities do not consistently improve performance across backbone--system combinations. Together, these observations suggest that effective recovery requires agents to use error feedback to revise assumptions and adjust subsequent actions, while additional turns alone do not ensure success.

\subsection{Capability Bottlenecks and Interactive Affordances (RQ3)}
\label{sec:capability_bottlenecks}

\textbf{The main bottleneck is maintaining executable evidence goals across stages.} Figure~\ref{fig:exp}(B4) shows a clear gap between identifying relevant evidence and carrying it through to mechanism-level reasoning: agents often know which modalities to inspect, yet fail to convert missing indicators into concrete follow-up actions. This reflects three coupled capability requirements: (1) maintaining which diagnostic subgoals remain unresolved, (2) preserving valid data and execution state across tool calls, and (3) integrating intermediate evidence into a consistent downstream explanation. The atomic profiles in Fig.~\ref{fig:exp}(B6) reinforce this view: relatively strong quantification or temporal analysis does not compensate for weaknesses in localization, visual interpretation, or evidence grounding. End-to-end diagnosis therefore depends on composing these capabilities rather than excelling at any single operation.

\textbf{Model scale does not directly determine realizable workflow capability.} Qwen3.8-27B illustrates how external feedback can compensate for weaker base capability. Although it matches several stronger models on Atomic PR in Table~\ref{tab:main_backbone_results}, its workflow performance varies substantially across systems in Table~\ref{tab:agent_system_results}, and it even surpasses Qwen3.8-Max under LAMBDA. Successful trajectories show that executable feedback---such as array shapes, missing-value patterns, and tool errors---helps the model revise local assumptions and preserve partial evidence. Conversely, stronger models can still fail when they do not convert recognized uncertainty into further analysis. GPT-5.6-Sol, for example, often identifies missing evidence and states limitations correctly, yet does not always turn these gaps into targeted follow-up computations. This contrast suggests that interactive execution primarily expands \emph{realizable} capability by improving evidence acquisition and control, rather than simply reflecting model size.

\textbf{Supplementary component ablations expose where execution capability is lost.} Figure~\ref{fig:exp}(B1) shows that removing code reduces Workflow Score from 28.5\% to 13.7\%; disabling skill routing in the no-code setting further lowers it to 12.4\%, while removing domain tools reduces performance to zero. These components support different parts of the workflow: code enables finer-grained spatial and multimodal analysis, skill routing helps select and invoke relevant procedural guidance, and domain tools provide direct access to standardized evidence. The result indicates that analytical intent alone is insufficient unless the agent can translate it into executable evidence.

%% file: tables_tex/tab_main_results.tex
\begin{table}[!t]
    \centering
    \caption{Main backbone results on DUDA-Bench under the ReAct system.}
    \label{tab:main_backbone_results}
    \small
    \setlength{\tabcolsep}{4.2pt}
    \renewcommand{\arraystretch}{1.02}

    \begin{tabular}{@{}lccccccc@{}}
        \toprule
        \multirow{2}{*}{\textbf{Model}}
        & \multirow{2}{*}{\makecell{\textbf{Atomic}\\\textbf{PR}}}
        & \multicolumn{4}{c}{\textbf{Workflow Outcome}}
        & \multicolumn{2}{c}{\textbf{Efficiency}} \\
        \cmidrule(lr){3-6}
        \cmidrule(lr){7-8}
        &
        & \textbf{Recognition}
        & \textbf{Localization}
        & \makecell{\textbf{Indicator}\\\textbf{F1}}
        & \makecell{\textbf{Workflow}\\\textbf{Score}}
        & \makecell{\textbf{Avg.}\\\textbf{Turns}}
        & \makecell{\textbf{Tokens}\\\textbf{(K)}} \\
        \midrule

        GPT-5.6-Terra
        & 0.419 & 0.250 & 0.080 & 0.063 & 0.071
        & \textbf{9.9} & \underline{184.7} \\

        Gemini-3.7-Flash
        & \textbf{0.558} & 0.432 & \underline{0.169}
        & \underline{0.195} & \underline{0.182}
        & 15.9 & 239.1 \\

        Claude-Sonnet-5
        & \textbf{0.558} & 0.386 & 0.159 & 0.111 & 0.135
        & 18.2 & 228.5 \\

        GPT-5.6-Sol
        & \underline{0.500} & 0.318 & 0.080 & 0.103 & 0.091
        & \underline{12.1} & \textbf{177.9} \\

        Kimi-K3
        & 0.465 & \underline{0.591} & 0.080 & 0.189 & 0.134
        & 13.4 & 196.8 \\

        Qwen3.8-Max
        & \underline{0.500} & \textbf{0.727} & \textbf{0.274}
        & \textbf{0.281} & \textbf{0.277}
        & 16.7 & 244.5 \\

        Qwen3.8-27B
        & \underline{0.500} & 0.432 & 0.125 & 0.141 & 0.133
        & 20.3 & 322.3 \\

        \bottomrule
    \end{tabular}

    \begin{minipage}{0.98\linewidth}
        \footnotesize
        \textit{Notes.}
        Atomic PR and Workflow PR denote the corresponding task-level performance metrics.
        Avg. Turns and Tokens (K) report the mean number of agent turns and total token usage
        (in thousands) per workflow episode. Repetitions are averaged within each problem,
        and problems are equally weighted. Best and second-best distinct results are shown
        in \textbf{bold} and \underline{underlined}, respectively; lower is better for
        efficiency metrics.
    \end{minipage}
\end{table}

%% file: tables_tex/tab_agent_comparison.tex
\begin{table}[!t]
    \centering
    \caption{Comparison of agent systems on DUDA-Bench using Qwen3.8-Max and Qwen3.8-27B backbones.}
    \label{tab:agent_system_results}
    \small
    \setlength{\tabcolsep}{3.5pt}
    \renewcommand{\arraystretch}{1.02}

    \begin{tabular}{@{}lcccc@{}}
        \toprule
        \textbf{Agent System}
        & \textbf{Recognition}
        & \textbf{Localization}
        & \textbf{Indicator F1}
        & \textbf{Workflow Score} \\
        \midrule

        \rowcolor{gray!10}
        \multicolumn{5}{l}{\textbf{\textit{Qwen3.8-Max}}} \\

        Direct
        & \underline{0.659}
        & \underline{0.216}
        & 0.126
        & \underline{0.171} \\

        Plan-and-Execute
        & 0.591
        & 0.125
        & \underline{0.128}
        & 0.126 \\

        ReAct
        & \textbf{0.727}
        & \textbf{0.274}
        & \textbf{0.281}
        & \textbf{0.277} \\

        MAF-Magentic
        & 0.295
        & 0.057
        & 0.081
        & 0.069 \\

        LAMBDA
        & 0.227
        & 0.125
        & 0.070
        & 0.097 \\

        \midrule

        \rowcolor{gray!10}
        \multicolumn{5}{l}{\textbf{\textit{Qwen3.8-27B}}} \\

        Direct
        & \textbf{0.500}
        & \underline{0.148}
        & 0.101
        & 0.124 \\

        Plan-and-Execute
        & 0.364
        & 0.091
        & 0.087
        & 0.089 \\

        ReAct
        & \underline{0.432}
        & 0.125
        & \underline{0.141}
        & \underline{0.133} \\

        MAF-Magentic
        & \underline{0.432}
        & 0.091
        & 0.096
        & 0.093 \\

        LAMBDA
        & \textbf{0.500}
        & \textbf{0.181}
        & \textbf{0.149}
        & \textbf{0.165} \\

        \bottomrule
    \end{tabular}

    \begin{minipage}{0.92\linewidth}
        \footnotesize
        \textit{Notes.}
        Repetitions are averaged within each problem, and problems are equally weighted.
        Best and second-best distinct results within each backbone are shown in
        \textbf{bold} and \underline{underlined}, respectively.
    \end{minipage}
\end{table}

%% file: figures_tex/fig_exp.tex
\begin{figure}[t]
    \centering
    \includegraphics[
        width=\columnwidth
    ]{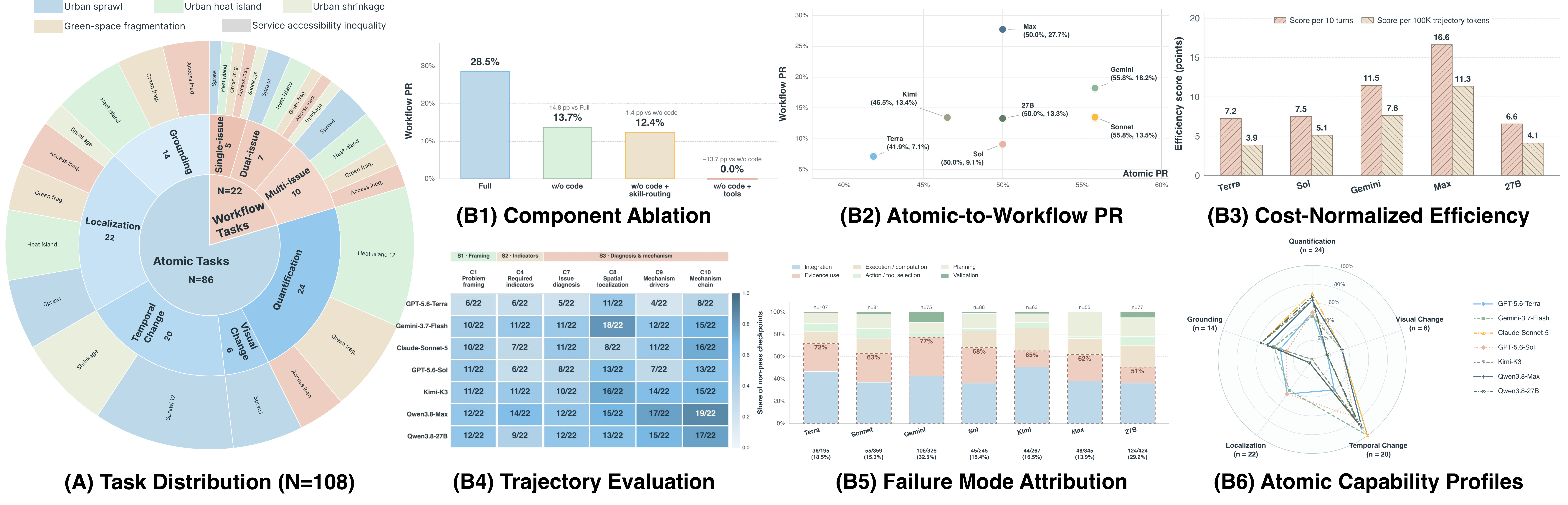}
    \caption{Benchmark composition and experimental analyses. (A) Task distribution across workflow and atomic tasks. (B1--B6) Component ablation, atomic-to-workflow performance, cost-normalized efficiency, trajectory evaluation, failure-mode attribution, and atomic capability profiles.}
    \label{fig:exp}
\end{figure}

%% file: sections/5-conclusion.tex
We introduce \textbf{DUDA-Bench}, a benchmark that formulates data-driven urban diagnosis as a unified agentic task spanning problem framing, urban profiling, diagnostic reasoning, and validation. DUDA-Bench combines hierarchical atomic and workflow tasks, a multimodal urban dataset, an interactive agentic environment, and outcome- and process-level evaluation to assess complete analytical workflows. Experiments across multiple backbone models and agent systems reveal a substantial gap between isolated analytical competence and end-to-end diagnosis, with recurring failures in evidence acquisition, state maintenance, recovery and cross-stage integration. At the same time, trajectory and component analyses show that verifiable interaction, code execution, skill routing, and domain tools can help agents better realize their analytical capabilities. We hope DUDA-Bench provides a useful foundation for evaluating and improving multimodal agents toward more reliable, evidence-grounded urban analysis.

%% file: sections/appendix.tex
\section{Benchmark Construction Details}
\label{app:benchmark-construction}

\subsection{Task Construction}
\label{app:task}
\input{figures_tex/fig_motivation}
\paragraph{Task illustration.} Fig.~\ref{fig:motivation} contrasts a predefined urban indicator estimation task with a DUDA urban diagnosis task. While the existing tasks ask the agent to perform a specified calculation, urban diagnosis requires it to inspect spatial patterns, select analytical steps, integrate tool outputs, and produce a localized, evidence-grounded diagnosis.

\paragraph{Task taxonomy.} Table~\ref{tab:task_statistics} summarizes the composition of DUDA-Bench: 22 workflow tasks categorized by problem complexity and urban problem type, and 86 atomic tasks categorized by capability type and urban problem type.
\input{tables_tex/tab_task_statistics}
\subsection{Detailed Statistics and Collection of dataset}
\label{app:dataset-statistics}

\input{tables_tex/tab3_dataset_statistics}
\paragraph{Dataset coverage.} Table~\ref{tab:dataset_statistics} summarizes the dataset's 12 city settings and 22 study-region cases, with observations spanning 2015--2024. Cities includes Atlanta, Detroit, Mumbai, Phoenix, Beijing, Delhi, Guangzhou, Kunming, Łódź, Shanghai, Valencia and Seoul. The dataset contains 7,071 data assets and 1,507,693 records across five modalities. Table~\ref{tab:data_source_details} details each source's analytical use, spatial level, temporal coverage, resolution, and access point.
\input{tables_tex/tab_app_data_source}

\subsection{Extended evaluation details and Human Verification}
\label{app:human-verification}
\input{tables_tex/tab_ckpt_rubric}
\paragraph{Trajectory Evaluation Details.} Table~\ref{tab:human_checkpoint_rubric} defines the 14 checkpoints used by both the LLM-as-a-Judge and human annotators to evaluate agent trajectories. Each applicable checkpoint is scored as Pass (1.0), Partial (0.5), or Fail (0.0) based on its completion and support from the recorded trajectory.

\input{tables_tex/tab_judge_alignment}

\paragraph{Human Verification.} We randomly sampled 56 ReAct workflow trajectories, with eight trajectories from each of the seven backbone models. Each trajectory was converted into an annotation packet containing exactly the information provided to the checkpoint judge: the task specification, checkpoint definitions with reference content, the compacted observable trajectory, and the final output, while the judge verdicts were withheld. Annotators A and B independently labelled every applicable checkpoint as \textit{pass}, \textit{partial}, or \textit{fail} using the same written guideline, which restates the judge rubric and specifies additional operational rules. Both annotators were blind to each other's annotations and to the judge outputs. Disagreements were resolved by a third reviewer applying the same guideline, producing the adjudicated human gold standard. As shown in Table~\ref{tab:judge_alignment}, A and B agreed on 93.6\% of the 716 checkpoint verdicts, corresponding to a 6.4\% disagreement rate. Annotator A and GLM-5.3 agreed on 89.8\% of checkpoint labels. Against the adjudicated gold standard, Annotator A reached 97.8\% accuracy, while GLM-5.3 achieved 90.4\%, a gap of 7.4 percentage points. When \textit{partial} and \textit{fail} are merged into a single non-pass class, the judge accuracy increases to 94.1\%, indicating that most remaining errors arise from distinguishing partial completion from complete failure rather than from incorrectly identifying successful checkpoints.

\subsection{Domain Skills}
\label{app:domain_skills}
DUDA-Env provides domain skills that guide agents in discovering case data, computing urban indicators, querying map features, inspecting imagery, and localizing spatial patterns. Table~\ref{tab:domain_skills_tools} summarizes these skills alongside representative executable operations. A geospatial sandbox additionally supports task-specific analyses beyond the predefined operations.
\input{tables_tex/tab_app_skill}
\section{Extended Related Work}
\label{app:extended-related-work}
\paragraph{Data-Driven Urban Diagnosis.} Urban problem diagnosis is an established research area in urban studies. Broad reviews characterize diagnosis as a basis for understanding urban conditions and informing appropriate interventions through historical, conceptual, and methodological perspectives~\cite{su10093306,BIBRI2021100047,FOSTER201692}. Empirical research further operationalizes this perspective through data-driven analyses tailored to individual urban problems and regions. Representative examples include urban sprawl and land fragmentation in Phoenix~\cite{SHRESTHA2012522}, urban shrinkage and socio-spatial inequality in Detroit~\cite{XIE2018350}, surface urban heat island patterns in Valencia~\cite{WEI2024103875}, and accessibility in Seoul~\cite{ijgi14070247}. Together, these studies demonstrate how heterogeneous spatial, temporal, and socioeconomic evidence can support urban diagnosis. However, such analyses are generally formulated around individual problem settings rather than as a shared computational task spanning multiple urban issues, regions, and data sources. DUDA-Bench builds on this literature by abstracting their common analytical structure into a unified agent-oriented formulation.

\paragraph{Benchmarks for Urban and Geospatial Intelligence.} Recent benchmarks have expanded urban intelligence evaluation along several complementary directions. Interactive benchmarks evaluate agents operating within urban environments: CityBench combines simulator-based understanding and decision making~\cite{10.1145/3711896.3737375}, CityEQA studies embodied question answering through city exploration~\cite{zhao-etal-2025-cityeqa}, and USTBench evaluates spatiotemporal reasoning at both process and outcome levels~\cite{lai2026ustbench}. A second line focuses on multimodal urban understanding. UrBench evaluates satellite, street-view, and cross-view reasoning~\cite{10.1609/aaai.v39i10.33163}, UBench integrates spatial, temporal, and visual urban information~\cite{Feng_2025_ICCV}, DynamicVL targets multi-temporal urban change~\cite{xuan2026dynamicvl}, and CityLens connects urban imagery with socioeconomic indicators~\cite{liu2026citylens}. Related geospatial and Earth-observation benchmarks extend this scope with multi-sensor vision~\cite{Danish_2025_ICCV}, agentic EO workflows~\cite{feng2026earthagent}, and executable GIS analysis~\cite{yu2026geoagentbenchdynamicexecutionbenchmark}. These benchmarks provide rich urban and geospatial evaluation settings, but their primary targets are individual perception, reasoning, prediction, navigation, or GIS capabilities. DUDA-Bench instead treats urban problem diagnosis itself as the evaluation target and requires multiple capabilities to be composed within a shared multi-stage workflow.

\paragraph{LLM Agent Benchmarks.} LLM agent evaluation has increasingly shifted from final-answer accuracy toward interactive execution and trajectory-level reliability. In general settings, $\tau$-bench evaluates consistent success in tool-mediated interactions using pass$^k$~\cite{yao2025taubench}; GTA-2 connects atomic tool use with open-ended workflows through checkpoint-based evaluation~\cite{wang2026gta2benchmarkinggeneraltool}; and TRAJECT-Bench evaluates tool selection, parameter grounding, and ordering along execution trajectories~\cite{he2026trajectbencha}. Scientific-agent benchmarks pursue a similar direction: SciAgentBench spans elementary actions and multi-tool workflows~\cite{shen2026sciagentgym}, MoSciBench evaluates coding agents for multimodal data-driven discovery~\cite{liu2026towards}, and ZephyrusBench grounds agent evaluation in programmatic weather analysis~\cite{varambally2026zephyrus}. DUDA-Bench follows this process-aware evaluation paradigm while extending it to multimodal urban diagnosis, where agents must connect problem framing, heterogeneous evidence acquisition, diagnostic reasoning, and validation within an end-to-end analytical process.

\section{Agent System Environment Configuration}
\label{app:prompt-details}

We instantiate five representative agent systems under a shared benchmark
environment. All systems receive the same task input, public data card,
benchmark-provided data and tool interfaces, and evidence-grounding policy,
while preserving their system-specific execution and orchestration mechanisms.
Unless otherwise specified, agents are restricted to the provided case bundle
and cannot access external web resources.

\subsection{ReAct System Prompt}

\begin{promptbox}{ReAct System Prompt}

\textbf{Role.}
You are an expert urban-analysis agent. You answer questions about a city or
region by inspecting multimodal geospatial data and writing analysis code.

\medskip
\textbf{Output Protocol.}

At every step, output a SINGLE JSON object and no text outside the JSON.
Choose exactly one action:

\begin{verbatim}
{"thought": "...", "action": "run_code",
 "code": "..."}
\end{verbatim}

\begin{verbatim}
{"thought": "...", "action": "call_tool",
 "tool_name": "...", "tool_args": {...}}
\end{verbatim}

\begin{verbatim}
{"thought": "...", "action": "finish",
 "final_answer": "..."}
\end{verbatim}

Execute at most one environment action at each interaction step.
After receiving the corresponding observation, decide the next action based on
the updated evidence.

\medskip
\textbf{Constraints.}

\begin{enumerate}[leftmargin=*, nosep]
    \item Use only the provided benchmark data and environment. Do not browse
    the web or access files outside the supplied case bundle.

    \item Ground substantive findings in observable evidence. Identify the
    corresponding data source, modality, artifact, or path when available.

    \item If the available evidence is insufficient, state the limitation
    explicitly instead of guessing.

    \item The complete current AOI data card is already provided below.
    Inspect it directly and do not invoke tools merely to rediscover the same
    available modalities.

    \item There is no fixed priority among numeric, visual, textual, and
    code-based evidence. Select actions according to the task and the
    available modalities.

    \item Qualitative judgments directly observable from RGB imagery
    (e.g., denser built-up areas, greener land cover, or visible land-cover
    change) are valid visual evidence. Thermal conditions and exact
    quantitative indices must be derived from the corresponding canonical
    raster, summary, or analytical data rather than inferred visually.

    \item Do not inspect Python source code, module dictionaries, configuration
    files, settings, or unrelated directory listings as task evidence.

    \item Follow the exact output format requested by the task.
\end{enumerate}

\medskip
\textbf{Execution Support.}

The agent additionally receives the benchmark sandbox API description used for
code execution. The full configuration also exposes the benchmark-provided
domain-skill instructions and available skill directories. These components
constitute the full ReAct system evaluated in the main experiments.

\medskip
\textbf{Public Data Card.}

\begin{verbatim}
<public_data_card>
{public_data_card}
</public_data_card>
\end{verbatim}

\end{promptbox}

\subsection{Direct Tool Use System Prompt}

\begin{promptbox}{Direct Tool Use System Prompt}

\textbf{Role.}
You are an expert urban-analysis assistant. Solve the task directly, without
using a fixed multi-step agent framework, iterative think--act--observe loop,
or explicit planning stage. You may use the provided environment actions for
one data-gathering round before answering, or answer immediately if the
provided evidence already suffices.

\medskip
\textbf{Constraints.}

\begin{enumerate}[leftmargin=*, nosep]
    \item You have at most ONE round of environment actions. All required
    actions must be selected before observing any tool result.

    \item Tool results cannot be used to generate additional actions or modify
    the arguments of previously selected actions. After all selected actions
    are executed, you must produce the final answer.

    \item Use only the provided benchmark data and environment. Do not browse
    the web or access files outside the supplied case bundle.

    \item Ground substantive findings in observable evidence. Identify the
    corresponding data source, modality, artifact, or path when available.

    \item If the available evidence is insufficient, state
    ``insufficient evidence'' instead of guessing.

    \item Qualitative judgments directly observable from RGB imagery
    (e.g., denser built-up areas, greener land cover, or visible land-cover
    change) are valid visual evidence. Thermal conditions and exact
    quantitative indices must be derived from the corresponding canonical
    raster, summary, or analytical data rather than inferred visually.

    \item The supplied public data card describes the available case data and
    should not be rediscovered through redundant environment calls.

    \item Follow the exact output format requested by the task.
\end{enumerate}

\medskip
\textbf{Output Protocol.}

The FIRST reply must be a SINGLE JSON object and contain no text outside the
JSON. Choose one of the following forms:

\begin{verbatim}
{"thought": "...",
 "tool_calls": [
   {"tool_name": "...", "tool_args": {...}},
   ...
 ]}
\end{verbatim}

or

\begin{verbatim}
{"thought": "...",
 "final_answer": "..."}
\end{verbatim}

Request all required actions in the single \texttt{tool\_calls} list. The
runtime may truncate the list according to the configured
environment-action budget.

\medskip
\textbf{Public Data Card.}

\begin{verbatim}
<public_data_card>
{public_data_card}
</public_data_card>
\end{verbatim}

\end{promptbox}

\subsection{Plan-and-Execute System Prompt}

\begin{promptbox}{Plan-and-Execute System Prompt}

\textbf{Planning Role.}
You are an urban-analysis planning agent. Given a question and the available
environment actions with their input schemas, produce a short, ordered plan
whose execution gathers sufficient evidence to answer the question.

The complete plan must be finalized before any environment observation is
seen. During execution, observations may be recorded for final synthesis but
must not alter, reorder, add, or remove planned actions.

\medskip
\textbf{Planning Constraints.}

\begin{enumerate}[leftmargin=*, nosep]
    \item Produce an ordered plan of environment actions, including tool calls
    or code execution when available.

    \item You will not observe action results before finalizing the complete
    plan and cannot replan afterward.

    \item Use only action names provided by the benchmark environment and
    supply all fields required by their input schemas.

    \item Use only the provided benchmark data and environment. Do not browse
    the web or access files outside the supplied case bundle.

    \item The public data card already describes the available case data; do
    not plan redundant actions merely to rediscover available modalities.

    \item Use at most \texttt{\{max\_steps\}} planned environment actions.
\end{enumerate}

\medskip
\textbf{Planning Output Protocol.}

Respond with a SINGLE JSON object and no text outside the JSON:

\begin{verbatim}
{
  "plan": [
    {
      "step": 1,
      "tool_name": "...",
      "tool_args": {...},
      "reason": "..."
    }
  ]
}
\end{verbatim}

\medskip
\textbf{Synthesis Role.}

You are an urban-analysis agent. The previously generated fixed plan has
already been executed. Answer the original question using ONLY the evidence
obtained from those actions.

\medskip
\textbf{Synthesis Constraints.}

\begin{enumerate}[leftmargin=*, nosep]
    \item Ground substantive claims in the actually observed evidence and
    identify the corresponding data source, modality, artifact, or path when
    available.

    \item If an action failed or the available evidence is insufficient, state
    the unresolved limitation rather than inventing additional evidence.

    \item Do not request further environment actions or revise the original
    plan.

    \item Follow the exact output format requested by the task.
\end{enumerate}

\medskip
\textbf{Public Data Card.}

\begin{verbatim}
<public_data_card>
{public_data_card}
</public_data_card>
\end{verbatim}

\end{promptbox}

\subsection{MAF System Prompts}

We instantiate the Microsoft Agent Framework (MAF) baseline using a
manager--analyst--reviewer organization. All three roles use the same backbone
model. Only the Analyst is allowed to execute benchmark environment actions;
the Reviewer performs read-only verification, while the Manager coordinates
the workflow and produces the final response.

\begin{promptbox}{MAF Analyst, Reviewer, and Manager Prompts}

\textbf{Common Policy.}

Use only the provided benchmark data and environment. Do not browse the web or
access files outside the supplied case bundle. Ground substantive findings in
observable evidence and identify the corresponding data source, modality,
artifact, or path when available. If evidence is insufficient, state the
limitation rather than guessing. Follow the exact output format requested by
the task.

\medskip
\textbf{Analyst.}

You are the urban-data Analyst and the only role allowed to perform analytical
environment actions.

You may invoke \texttt{dura\_run\_code} and
\texttt{dura\_call\_tool}. Select the appropriate interface according to the
analytical need and available evidence. Inspect returned observations and
revise or debug the analysis when necessary. Use environment actions
efficiently and avoid redundant calls.

Do not inspect Python source code, module dictionaries, settings,
configuration files, or unrelated directory listings; these are not task
evidence. Once the assigned evidence has been collected, return the evidence
and any unresolved limitations to the Manager rather than continuing
environment discovery.

Across the entire case, code execution and domain or visual tools share a
budget of \texttt{\{max\_environment\_actions\}} executed environment actions.
If a tool reports an \texttt{ANALYST\_*\_BUDGET\_EXHAUSTED} condition,
immediately return the best available evidence and unresolved limitations and
do not retry the exhausted action.

Available DURA tool schemas are provided by the runtime.

\medskip
\textbf{Reviewer.}

You are a read-only evidence and answer Reviewer and have no environment
execution tools.

Review the Analyst's evidence, calculations, citations, and coverage. Identify
specific substantive gaps, unsupported claims, or analytical inconsistencies.
Do not invent observations or request purely cosmetic revisions.

If the evidence and current answer substantively satisfy the original request,
begin with \texttt{APPROVED}. Otherwise, begin with
\texttt{NEEDS\_REVISION} and list only blocking corrections.

\medskip
\textbf{Manager.}

You are the Magentic task Manager. Coordinate the Analyst and Reviewer and
produce the final answer.

Plan and coordinate the participants, but do not perform analysis yourself and
do not claim observations that the participants did not produce. Treat
internal planning and progress-ledger requests as orchestration protocol
messages rather than benchmark answers.

When the benchmark requests JSON, return one complete and compact JSON object
with no markdown or additional text.

\medskip
\textbf{Public Data Card.}

\begin{verbatim}
<public_data_card>
{public_data_card}
</public_data_card>
\end{verbatim}
\end{promptbox}

\subsection{LAMBDA System Prompt}

For LAMBDA, we use a lightweight benchmark-facing adaptation rather than its
original interactive interface. Following the sequential tool-execution setup
used for data-science-agent evaluation in EvoDS~\cite{10.1145/3770855.3818002}, the agent
iteratively selects an analytical function, observes its actual execution
result, and adapts subsequent actions. This preserves LAMBDA's
data-analysis-oriented iterative execution while replacing external execution
components with the common DURA environment for controlled comparison.

\begin{promptbox}{LAMBDA System Prompt}

\textbf{Role.}

You are a LAMBDA-style data-analysis agent operating on a DURA-Bench case.
Solve the task by repeatedly selecting a native analytical function,
inspecting its real observation, and adapting the next step. When sufficient
evidence has been collected, return the final answer as ordinary text and make
no further function call.

\medskip
\textbf{Execution Policy.}

\begin{enumerate}[leftmargin=*, nosep]
    \item Use only the supplied functions and public AOI bundle. Never browse
    the web, invoke an external shell, access host files, or invent
    observations.

    \item Use environment actions efficiently and avoid redundant calls. Do
    not repeat an unchanged failed or successful action. The hard limit is
    \texttt{\{max\_environment\_actions\}} environment actions.

    \item \texttt{execute\_python} provides a persistent analysis sandbox:
    serializable variables and data frames may persist across execution cells,
    and common imports are refreshed automatically when needed.

    \item Inspect returned observations before deciding the next analytical
    action and revise the analysis when execution results indicate an error or
    inconsistency.

    \item Ground substantive findings in observable evidence and identify the
    data paths, modalities, and years actually inspected when available.

    \item If the available evidence is insufficient, state the limitation
    explicitly rather than guessing.

    \item Visual tool outputs may contain actual image pixels. Use them for
    qualitative judgments directly supported by the image, but do not infer
    exact quantitative raster values or thermal conditions merely by visual
    inspection.

    \item Do not inspect source code, configuration files, module dictionaries,
    or unrelated directory listings as analytical evidence.

    \item Follow the exact output format requested by the task.
\end{enumerate}

\medskip
\textbf{Workflow Protocol.}

The runtime supplies the benchmark-specific workflow protocol describing the
available native functions, their schemas, execution rules, and termination
conditions.

\medskip
\textbf{Public Data Card.}

\begin{verbatim}
<public_data_card>
{public_data_card}
</public_data_card>
\end{verbatim}

\end{promptbox}

%% file: figures_tex/fig_motivation.tex
\begin{figure}[H]
    \centering
    \includegraphics[
        width=\columnwidth
    ]{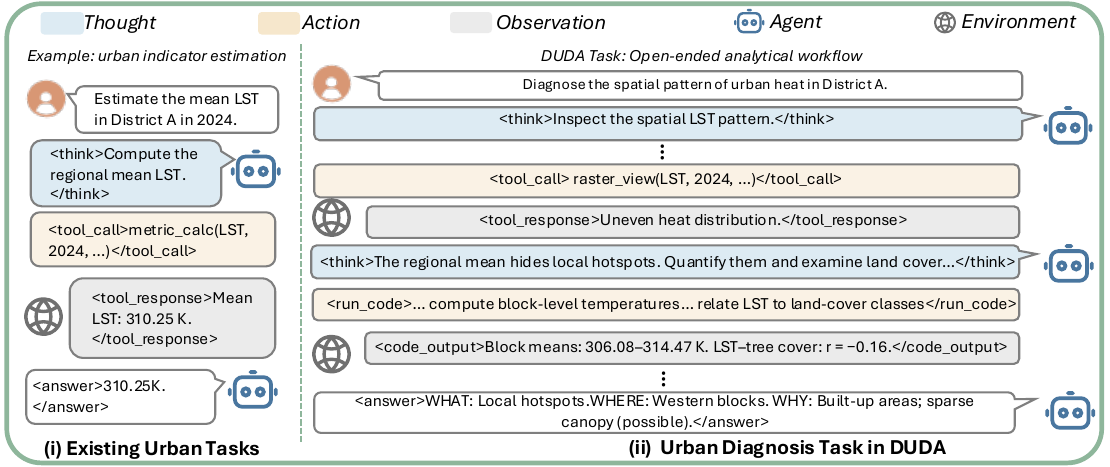}
    \caption{From predefined urban analysis to open-ended urban diagnosis. Existing urban tasks typically evaluate a specified analytical operation, whereas DUDA requires an agent to iteratively determine, acquire, and integrate evidence throughout a diagnostic workflow.}
    \label{fig:motivation}
\end{figure}

%% file: tables_tex/tab_task_statistics.tex
\begin{table}[H]
    \centering
    \caption{Task composition and distribution of DUDA-Bench.}
    \label{tab:task_statistics}

    \small
    \setlength{\tabcolsep}{3pt}
    \renewcommand{\arraystretch}{1.08}

    \begin{tabularx}{\columnwidth}{
        @{}
        >{\raggedright\arraybackslash}p{0.34\columnwidth}
        >{\centering\arraybackslash}p{0.045\columnwidth}
        >{\raggedright\arraybackslash}X
        >{\centering\arraybackslash}p{0.045\columnwidth}
        @{}
    }
        \toprule
    
        \multicolumn{2}{c}{
            \textbf{Workflow Tasks} (\(N=22\))
        }
        &
        \multicolumn{2}{c}{
            \textbf{Atomic Tasks} (\(N=86\))
        } \\
    
        \midrule
    
        \rowcolor{gray!10}
        \mbox{\textbf{\textit{Problem Complexity (3)}}}
        & \textbf{\textit{N}}
        & \mbox{\textbf{\textit{Capability Types (5)}}}
        & \textbf{\textit{N}} \\
    
        Single-issue
        & 5
        & Indicator quantification
        & 32 \\
    
        Dual-issue
        & 7
        & Temporal change analysis
        & 20 \\
    
        Multi-issue
        & 10
        & Spatial problem localization
        & 22 \\
    
        &
        &
        Visual change interpretation
        & 6 \\
    
        &
        &
        Evidence-to-metric grounding
        & 6 \\
    
        \bottomrule
    \end{tabularx}

    \vspace{4pt}

    \begin{tabularx}{\columnwidth}{
        @{}
        >{\raggedright\arraybackslash}X
        >{\centering\arraybackslash}p{0.18\columnwidth}
        >{\centering\arraybackslash}p{0.18\columnwidth}
        @{}
    }
        \toprule

        \rowcolor{gray!10}
        \textbf{\textit{Problem Types (5)}}
        & \textbf{\textit{Workflow N}}
        & \textbf{\textit{Atomic N}} \\

        Urban Morphology: Urban sprawl
        & 6
        & 24 \\

        Eco-Climatic Resilience: Urban heat island
        & 6
        & 24 \\

        Eco-Climatic Resilience: Green-space fragmentation
        & 4
        & 16 \\

        Service accessibility inequality
        & 4
        & 12 \\

        Urban Morphology: Urban shrinkage
        & 2
        & 10 \\

        \bottomrule
    \end{tabularx}
\end{table}

%% file: tables_tex/tab3_dataset_statistics.tex
\begin{table}[t]
    \centering
    \caption{Statistics and data-family composition of the multisource urban case dataset.}
    \label{tab:dataset_statistics}
    \small
    \setlength{\tabcolsep}{4pt}
    \renewcommand{\arraystretch}{1.12}

    \begin{tabularx}{\columnwidth}{
        @{}
        >{\raggedright\arraybackslash}p{0.27\columnwidth}
        >{\centering\arraybackslash}p{0.17\columnwidth}
        >{\raggedright\arraybackslash}X
        @{}
    }
        \toprule
        \textbf{Aspect}
        & \textbf{Count}
        & \textbf{Coverage} \\
        \midrule

        \rowcolor{gray!10}
        \multicolumn{1}{@{}l}{
            \textbf{\textit{Spatial and temporal coverage}}
        }
        & \textbf{\textit{Count}}
        & \textbf{\textit{Range}} \\

        Cities / Regions
        & \textbf{12 / 22}
        & City / study-region cases \\

        Observation Years
        & \textbf{10}
        & 2015--2024 \\

        \midrule

        \rowcolor{gray!10}
        \multicolumn{1}{@{}l}{
            \textbf{\textit{Data composition}}
        }
        & \textbf{\textit{No. of types}}
        & \textbf{\textit{Included data}} \\

        Source Families
        & \textbf{8}
        & OSM; WorldPop; GHSL; VIIRS; Landsat; Sentinel-2; Mapillary; WorldCover \\

        Map
        & \textbf{2}
        & Roads; POIs/AOIs \\

        Indicator
        & \textbf{5}
        & Population; built-up areas; nighttime lights; Land Surface Temperature(LST); land cover \\

        Overhead
        & \textbf{1}
        & Satellite imagery \\

        Street
        & \textbf{1}
        & Street-view imagery \\

        Modalities
        & \textbf{5}
        & Remote sensing; street view; raster; vector; text \\

        \midrule

        \textbf{Total Data Assets}
        & \textbf{7,071}
        & - \\

        \midrule
        \textbf{Total Data Records }
        & \textbf{1,507,693}
        & - \\
        
        \bottomrule
    \end{tabularx}
\end{table}

%% file: tables_tex/tab_app_data_source.tex
\begin{table*}[t] \centering \caption{Data sources, contents, analysis scales, and spatial resolutions of the multimodal urban case dataset.} \label{tab:data_source_details} \small \setlength{\tabcolsep}{4pt} \renewcommand{\arraystretch}{1.10} \begin{tabularx}{\textwidth}{@{}>{\raggedright\arraybackslash}p{0.17\textwidth}>{\raggedright\arraybackslash}X>{\centering\arraybackslash}p{0.11\textwidth}>{\centering\arraybackslash}p{0.11\textwidth}>{\centering\arraybackslash}p{0.11\textwidth}@{}} \toprule \textbf{Source} & \textbf{Data Content and Benchmark Use} & \textbf{Analysis Scale} & \textbf{Resolution} & \textbf{Access} \\ \midrule OpenStreetMap & Administrative boundaries and vector features, including roads, buildings, land use, and amenities, for spatial delineation and accessibility analysis. & City/Region & -- & \href{https://www.openstreetmap.org/}{OSM} \\ WorldPop & Annual gridded population estimates for population change, density, and population-weighted accessibility analysis. & City/Region & 1 km & \href{https://data.worldpop.org/}{WorldPop} \\ VIIRS Annual VNL & Annual nighttime-light composites used as proxies for urban activity and its temporal changes. & City & 15 arcsec & \href{https://eogdata.mines.edu/products/vnl/}{EOG} \\ Landsat Collection 2 Level-2 & Surface-temperature rasters for thermal analysis, with Landsat 8 RGB imagery supplementing early periods without available Sentinel-2 imagery. & City/Region & 30 m grid & \href{https://planetarycomputer.microsoft.com/dataset/landsat-c2-l2}{PC catalog} \\ Sentinel-2 L2A & True-color satellite imagery for visual analysis of urban expansion, greenery, and surface change. & City/Region & 10 m (RGB) & \href{https://planetarycomputer.microsoft.com/dataset/sentinel-2-l2a}{PC catalog} \\ GHSL GHS-BUILT-S & Multi-epoch built-up-surface estimates for analyzing urban expansion, sprawl, and stagnation. & City & 100 m & \href{https://human-settlement.emergency.copernicus.eu/ghs_buS2023.php}{GHSL} \\ ESA WorldCover & Land-cover maps distinguishing tree cover, grassland, cropland, built-up areas, water, and other classes. & City/Region & 10 m & \href{https://esa-worldcover.org/en/data-access}{WorldCover} \\ Mapillary & Geotagged street-view images for inspecting street environments, facilities, building conditions, and visible greenery. & Region & -- & \href{https://www.mapillary.com/developer/api-documentation}{API docs} \\ \bottomrule \end{tabularx} \par\smallskip \begin{minipage}{\textwidth} \footnotesize \textit{Note.} Analysis scale denotes use within DUDA-Bench. Raster resolutions refer to the selected source products before benchmark-specific processing. Landsat 8/9 thermal observations have a native resolution of 100 m and are resampled to a 30 m product grid. Dashes indicate vector features or street-view photographs without a uniform ground sampling distance. PC denotes Microsoft Planetary Computer. \end{minipage} \end{table*}

%% file: tables_tex/tab_ckpt_rubric.tex
\begin{table*}[t]
\centering
\caption{Checkpoint rubric for trajectory-level evaluation.}
\label{tab:human_checkpoint_rubric}
\small
\setlength{\tabcolsep}{4pt}
\renewcommand{\arraystretch}{1.08}
\begin{tabularx}{\textwidth}{@{}c l >{\raggedright\arraybackslash}X@{}}
\toprule
ID & Checkpoint & Evaluation Criterion \\
\midrule
C1 & Issue Context & Identifies the urban issue and establishes an appropriate analytical context. \\
C2 & Spatiotemporal Representation & Establishes the relevant spatial scope, temporal scope, and analysis units. \\
C3 & Multimodal Evidence Planning & Selects or plans evidence sources that are relevant to the analytical objective. \\
C4 & Indicator Acquisition & Obtains or derives the indicators needed to characterize the urban condition. \\
C5 & Comparative Analysis & Performs appropriate temporal, spatial, or reference-based comparisons using the obtained evidence. \\
C6 & Multimodal Evidence Integration & Relates evidence across relevant modalities rather than treating observations in isolation. \\
C7 & Diagnostic Interpretation & Forms a diagnosis that is explicitly supported by the preceding observations and analyses. \\
C8 & Problem Localization & Localizes the diagnosed phenomenon to appropriate spatial areas when required by the task. \\
C9 & Mechanism Driver Analysis & Identifies plausible explanatory drivers grounded in observable evidence. \\
C10 & Mechanism Chain & Connects supported drivers and observed urban outcomes into a coherent diagnostic chain without unsupported causal claims. \\
C11 & Heterogeneity / Spillover & Examines relevant spatial heterogeneity, local variation, or spillover effects when applicable. \\
C12 & Validation and Uncertainty & Checks intermediate conclusions against available evidence and acknowledges relevant uncertainty or limitations. \\
C13 & Planning-Oriented Synthesis & Integrates the diagnosis and supporting evidence into planning-relevant conclusions or recommendations. \\
C14 & Data Validity Awareness & Recognizes material data-quality, coverage, temporal, or modality limitations that affect the analysis. \\
\bottomrule
\end{tabularx}
\par\smallskip
\begin{minipage}{\textwidth}
\footnotesize\textit{Scoring:} Pass (1.0): substantively completed and grounded in the trajectory; Partial (0.5): partially completed but incomplete or weakly grounded; Fail (0.0): absent, incorrect, or unsupported.
\end{minipage}
\end{table*}

%% file: tables_tex/tab_judge_alignment.tex

\begin{table}[t]
\centering
\caption{Agreement between human verification and the LLM-as-a-Judge (GLM-5.3) on trajectory checkpoints. Three-level labels (\textit{pass} / \textit{partial} / \textit{fail}); 716 checkpoint verdicts from 56 trajectories.}
\label{tab:judge_alignment}
\setlength{\tabcolsep}{10pt}

\begin{tabular}{lcc}
\toprule[1.2pt]
\textbf{Evaluation} & \textbf{Raw Agreement} & \textbf{Accuracy vs.\ Gold} \\
\midrule[0.6pt]
Annotator A vs.\ Annotator B & 93.6\% & -- \\
Annotator A vs.\ GLM-5.3 & 89.8\% & -- \\
\midrule[0.6pt]
Human annotator (A) & -- & 97.8\% \\
GLM-5.3 & -- & 90.4\% \\
Accuracy gap & -- & 7.4 pp \\
\bottomrule[1.2pt]
\end{tabular}
\end{table}

%% file: tables_tex/tab_app_skill.tex
\begin{table*}[t]
\centering
\caption{Domain skills and representative executable operations in DUDA-Env.}
\label{tab:domain_skills_tools}
\scriptsize
\setlength{\tabcolsep}{3.5pt}
\renewcommand{\arraystretch}{1.12}
\begin{tabularx}{\textwidth}{
    @{}
    >{\raggedright\arraybackslash}p{0.14\textwidth}
    >{\raggedright\arraybackslash}p{0.18\textwidth}
    >{\raggedright\arraybackslash}p{0.32\textwidth}
    >{\raggedright\arraybackslash}X
    @{}
}
\toprule
\textbf{Capability} & \textbf{Domain Skill} & \textbf{Representative Operations} & \textbf{Function and Output} \\
\midrule

Data discovery
& \texttt{data-card}
& \texttt{aoi\_data\_inventory}; \texttt{data\_access\_guide}
& Identifies available data sources, years, files, spatial scopes, and access methods for the current case. \\

Numerical analysis
& \texttt{urban-metrics}
& \texttt{annual\_mean\_metric}; \texttt{annual\_total\_metric}; \texttt{temporal\_change\_rate}; \texttt{temporal\_series}; \texttt{landcover\_class\_share}; \texttt{population\_weighted\_}\newline\texttt{facility\_distance}; \texttt{raster\_percentile}
& Computes annual statistics, temporal changes, land-cover shares, accessibility measures, and raster percentiles. \\

Map and facility queries
& \texttt{osm-map-query}
& \texttt{poi\_query}; \texttt{building\_query}; \texttt{road\_query}; \texttt{landuse\_query}
& Retrieves OSM roads, buildings, land-use polygons, and facilities such as hospitals, schools, and parks. \\

Satellite imagery access
& \texttt{raster-imagery}
& \texttt{temporal\_satellite\_}\newline\texttt{retrieval}; \texttt{regional\_satellite\_}\newline\texttt{retrieval}
& Selects Sentinel-2 imagery by year and region, using Landsat imagery when the required early observation is unavailable. \\

Raster visualization
& \texttt{raster-imagery}
& \texttt{raw\_raster\_visualization}; \texttt{normalized\_raster\_}\newline\texttt{visualization}; \texttt{thematic\_raster\_}\newline\texttt{visualization}
& Converts LST, population, built-up, nighttime-light, and land-cover rasters into interpretable maps and summary statistics. \\

Spatial localization
& \texttt{spatial-}\newline\texttt{localization}
& \texttt{aoi\_grid\_generation}; \texttt{region\_bbox\_resolution}
& Divides a city or study region into analysis units and resolves benchmark region identifiers to predefined spatial bounds. \\

Street-view inspection
& \texttt{street-view}
& \texttt{mapillary\_streetview\_}\newline\texttt{sampling}
& Samples geotagged Mapillary images and returns image files, locations, and capture metadata. \\

Custom geospatial analysis
& \texttt{geospatial-}\newline\texttt{sandbox}
& \texttt{custom\_geospatial\_}\newline\texttt{computation}
& Executes Python for raster statistics, spatial joins, aggregation, visualization, and other task-specific operations. \\

\bottomrule
\end{tabularx}
\end{table*}